\documentclass[sigconf]{aamas}

\usepackage{balance} 

\usepackage{graphics} 
\usepackage{epsfig} 
\usepackage{amsmath} 
\usepackage{amsfonts}
\usepackage{calligra}
\usepackage{xcolor}
\usepackage[caption=false,font=footnotesize]{subfig}

\usepackage{empheq}
\usepackage{hyperref}

\usepackage{booktabs}
\usepackage{makecell}
\usepackage{multirow}
\usepackage{array}
\newcolumntype{P}[1]{>{\centering\arraybackslash}p{#1}}
\usepackage{hyperref}

\newcommand{\cmt}[1]{{\color{black} #1}}

\newcommand{\norm}[1]{\left\lVert#1\right\rVert}

\makeatletter
\gdef\@copyrightpermission{
  \begin{minipage}{0.2\columnwidth}
   \href{https://creativecommons.org/licenses/by/4.0/}{\includegraphics[width=0.90\textwidth]{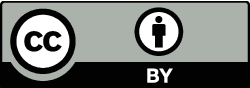}}
  \end{minipage}\hfill
  \begin{minipage}{0.8\columnwidth}
   \href{https://creativecommons.org/licenses/by/4.0/}{This work is licensed under a Creative Commons Attribution International 4.0 License.}
  \end{minipage}
  \vspace{5pt}
}
\makeatother

\setcopyright{ifaamas}
\acmConference[AAMAS '25]{Proc.\@ of the 24th International Conference
on Autonomous Agents and Multiagent Systems (AAMAS 2025)}{May 19 -- 23, 2025}
{Detroit, Michigan, USA}{Y.~Vorobeychik, S.~Das, A.~Nowé  (eds.)}
\copyrightyear{2025}
\acmYear{2025}
\acmDOI{}
\acmPrice{}
\acmISBN{}

\acmSubmissionID{<<fp1317>>}

\title[Ready, Bid, Go! On-Demand Delivery Using Fleets of Drones with Unknown, Heterogeneous Energy Storage Constraints]{Ready, Bid, Go! On-Demand Delivery Using Fleets of Drones with Unknown, Heterogeneous Energy Storage Constraints}

\author{Mohamed S. Talamali}
\affiliation{
  \institution{The University of Sheffield}
  \city{Sheffield}
  \country{United Kingdom}}
\email{m.s.talamali@sheffield.ac.uk}

\author{Genki Miyauchi}
\affiliation{
  \institution{The University of Sheffield}
  \city{Sheffield}
  \country{United Kingdom}}
\email{g.miyauchi@sheffield.ac.uk}

\author{Thomas Watteyne}
\affiliation{
  \institution{Inria} 
  \city{Paris}
  \country{France}}
\email{thomas.watteyne@inria.fr}

\author{Micael S. Couceiro}
\affiliation{
  \institution{Ingeniarius, Ltd.}
  \city{Alfena}
  \country{Portugal}}
\email{micael@ingeniarius.pt}

\author{Roderich Gro{\ss}}
\affiliation{
  \institution{
  Technical University of Darmstadt}
  \city{Darmstadt}
  \country{Germany}}
\email{roderich.gross@tu-darmstadt.de}

\begin{abstract}
Unmanned Aerial Vehicles (UAVs) are expected to transform logistics, reducing delivery time, costs, and emissions. This study addresses an on-demand delivery , in which fleets of UAVs are deployed to fulfil orders that arrive stochastically. Unlike previous work, it considers UAVs with heterogeneous, unknown energy storage capacities and assumes no knowledge of the energy consumption models. We propose a decentralised deployment strategy that combines auction-based task allocation with online learning. Each UAV independently decides whether to bid for orders based on its energy storage 
charge level, the parcel mass, and delivery distance. Over time, it refines its policy to bid only for orders within its capability. Simulations using realistic UAV energy models 
reveal that, counter-intuitively, assigning orders to the least confident bidders reduces delivery times and increases the number of successfully fulfilled orders. This strategy is shown to outperform threshold-based methods which require UAVs to exceed specific charge levels at deployment. We propose a variant of the strategy which uses learned policies for forecasting. This enables UAVs with insufficient charge levels to commit to fulfilling orders at specific future times, helping to prioritise early orders. Our work provides new insights into long-term deployment 
of UAV swarms, highlighting the advantages of decentralised energy-aware decision-making coupled with online learning in real-world dynamic environments.
\end{abstract}


\newcommand{\BibTeX}{\rm B\kern-.05em{\sc i\kern-.025em b}\kern-.08em\TeX}

\begin{document}




\pagestyle{fancy}
\fancyhead{}


\maketitle 


\section{Introduction}

Unmanned Aerial Vehicle (UAV)-based on-demand delivery is expected to bring significant benefits to society, including reductions in delivery times, operational costs, and carbon emissions. By deploying a fleet of UAVs, multiple orders can be fulfilled simultaneously, 
enabling large-scale applications in parcel, meal, and medication deliveries.

Various methodologies have been explored to address delivery problems with fleets of energy-constrained UAVs, including those that optimise order allocation and route planning. These include mixed-integer linear programming (MILP)~\cite{song2018persistent, agatz2018optimization, schermer2019hybrid, kitjacharoenchai2019multiple}, queuing theory~\cite{grippa2019drone}, auctions~\cite{RINALDI2022}, and various heuristic methods~\cite{dorling2016vehicle,peng2019hybrid,poikonen2020multi}.
An interesting problem variant is on-demand deliveries, where orders arrive stochastically over time~\cite{huang2021stochastic, liu2019optimization, chen2021improved}.

\begin{figure}[!t]
    \centering
    \includegraphics[width=0.95\linewidth]{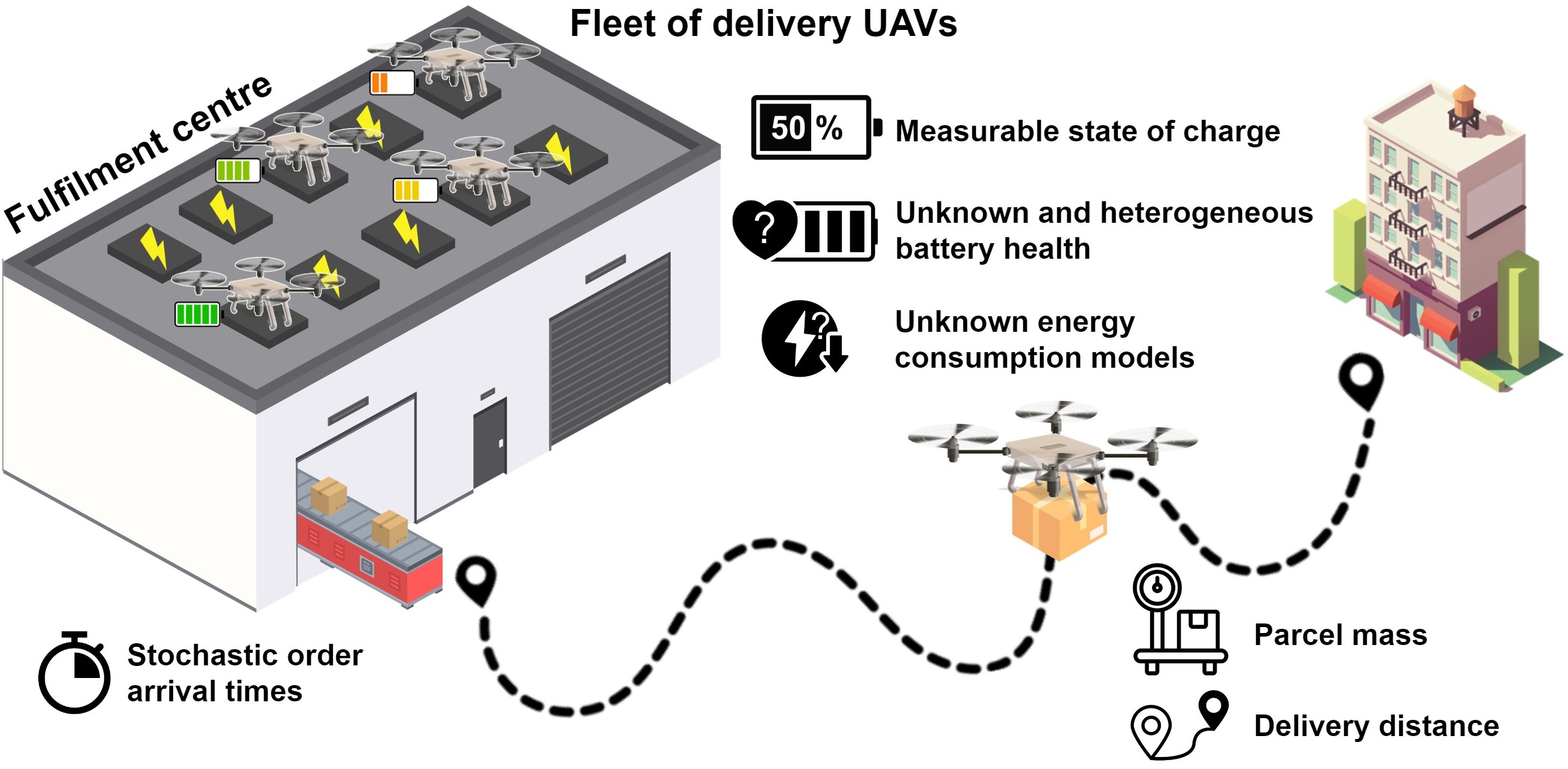}
    \Description[<Illustration of a UAV-based on-demand delivery scenario>]{<Illustration of a UAV-based on-demand delivery scenario>}
    \caption{On-demand delivery scenario with a fleet of UAVs committing to deliver orders arriving at a fulfilment centre. The fleet is heterogeneous, as the UAVs differ in battery health, and hence in their true energy storage capacities. Each UAV learns 
    a policy for placing bids on incoming orders based on its current charge level, parcel mass and delivery distance. Optionally, it can use this policy for forecasting, enabling it to plan the fulfilment of orders in the future.
    }
    \label{fig:illustration}
\end{figure}

Despite these advancements, studies often rely on several idealised assumptions that may not hold in real-world settings. First, they typically assume that the UAVs know their energy consumption models and can use them when planning their deliveries. This poses the risk of over-reliance on the given models: 
While various energy models have been proposed, it has been shown that different models can lead to divergent outcomes even under the same conditions~\cite{zhang2021energy}. Second, they commonly assume that UAVs are homogeneous. Yet in practice, even UAVs of the same make and model would differ due to variations in battery health, hardware wear, and other factors. These variations effectively result in a unique energy model for each UAV. Finally, they commonly assume the UAVs to charge instantaneously when visiting their base. This may hinder application in practical scenarios, as not reflecting the low duty cycles of most commercial delivery UAVs (e.g. 6.97\% for DJI FlyCart 30 with one battery). 
Factoring in the significant proportion of time devoted to charging ultimately leads to a new class of problems. For example, UAVs opting to dispatch with only a partial charge could result in faster deliveries than those waiting to charge fully.
The contributions of this paper are threefold:
\begin{enumerate}
    \item \textbf{Decentralised learning-based deployment strategy for UAVs with unknown battery health:} We propose a decentralised learning-based deployment strategy for on-demand delivery using UAV fleets. Each UAV decides whether to bid on incoming orders based on its battery’s \emph{state of charge}, parcel mass, and delivery distance. Unlike prior work, our approach requires no knowledge of the battery's \emph{state of health} (i.e., its true capacity) or the energy consumption model. Over time, each UAV learns to bid only on orders that it is likely to fulfil, enhancing the efficiency of the fleet.
    \item \textbf{Long-term evaluation of heterogeneous fleets:} We evaluate the proposed strategy using simulated UAV fleets that vary in battery health due to factors such as ageing, usage, and storage patterns. Simulations using realistic UAV energy models are conducted over a period of eight weeks with orders arriving stochastically at different rates. The results show that deploying the least confident bidding UAVs leads to the highest number of fulfilled orders and the lowest delivery times. Furthermore, the strategy outperforms traditional approaches that dispatch UAVs only when their battery charge exceeds predefined thresholds.
    \item \textbf{Forecasting-enabled order commitment:} We propose a second variant of the strategy where learned bidding policies are used for forecasting. This enables UAVs with insufficient charge levels to commit to fulfilling current orders at a specific future time, improving prioritisation of earlier orders and aligning better with first-come, first-served scheduling.
\end{enumerate}


The remainder of this paper is organised as follows. Section~\ref{sec:problem} presents the problem formulation. Section~\ref{sec:method} details the proposed decentralised learning-based deployment strategy. Sections~\ref{sec:experimental-setup} and~\ref{sec:results} present the simulator and results, respectively. Finally, Section~\ref{sec:conclusion} concludes the paper.

\section{PROBLEM STATEMENT}
\label{sec:problem}

As illustrated in Figure~\ref{fig:illustration}, a fulfilment centre (FC) operates a fleet of delivery UAVs, denoted as \( U = \{1, 2, \dots, S\} \). During an operational period of length \( T \), the FC stochastically receives orders \( O = \{1, 2, \dots\} \), which arrive at times \( t_1, t_2, \dots \), where \( 0 \leq t_1 < t_2 < \dots < T \). Each order \( j \in O \) corresponds to a delivery task involving a parcel of mass \( m_j \in [m_{\text{min}}, m_{\text{max}}] \), which is to be delivered to a location at a distance \( d_j \in [d_{\text{min}}, d_{\text{max}}] \) from the FC.

The FC maintains a communication infrastructure (e.g., a wireless network) to  
communicate with available UAVs (e.g., to announce delivery tasks) and allows UAVs to communicate with each other, facilitating collaborative decision-making.

All UAVs are powered by an on-board battery with a \emph{theoretical capacity} \(C_{\text{theoretical}}\). The battery of UAV \( i \in U \) has a \textit{state of health (SoH)}, denoted as \( \text{SoH}_i \in [0, 1] \), which determines its true (maximum) capacity. Specifically, the \emph{true battery capacity} is given by \(C_{\text{true},i}=\text{SoH}_i \cdot C_{\text{theoretical}} \). In this study, \( \text{SoH}_i \) is assumed to remain constant throughout the operational period.

At any given time \(t\), UAV \(i \in U\) possesses a \textit{state of charge (SoC)} denoted by \( \text{SoC}_i(t) \in [0, 100] \) which represents the percentage of \( C_{\text{true},i} \) that is currently available in its battery. At \(t = 0\), all UAVs are assumed to be fully charged, such that \( \forall i \in U: \text{SoC}_i(0) = 100 \), hence, each UAV stores $100\%$ of its true capacity. The change in the state of charge of UAV \( i \) depends on its activity:
\begin{itemize}
	\item \emph{Charging}: When within the FC, \( \text{SoC}_i \) increases at a rate of \( \lambda_{\text{charge}} \) per unit time until it reaches the maximum value.
	\item \emph{Delivering}: When outside the FC, \( \text{SoC}_i \) decreases at a rate of \( \lambda_{\text{delivery}}(m_p) \) per unit time, where \( m_p \) denotes its payload. While carrying a parcel of mass \( m \), we have \( m_p = m \); while not carrying a parcel, \( m_p = 0 \).
\end{itemize}
Each UAV can measure its state of charge ($\text{SoC}_i(t)$) at any time, but is unaware of its state of health ($\text{SoH}_i$), true capacity ($C_{\text{true},i}$) and energy consumption model (\( \lambda_{\text{delivery}}(m_p) \)).

When outside the FC, a UAV moves at a constant speed of \(v_a\). It first moves towards the parcel's delivery destination. Upon successful delivery, it immediately returns to the FC.\footnote{For simplicity, we assume UAVs can move directly towards their destinations. Our methods could be extended to accommodate UAVs navigating around static obstacles, provided a map of the environment is available.} While transporting a parcel to its delivery destination, a UAV may choose to abort the delivery and return to the FC. We refer to this as an \textit{aborted delivery attempt}. If a UAV's SoC reaches zero during delivery, it becomes unable to return to the FC and is considered lost.

The objective is to define a decentralised deployment strategy that assigns advertised delivery tasks to specific UAVs. Various performance criteria are explored, including (i) the number of successfully delivered parcels (the more, the better); (ii) the time elapsed from order arrival to successful delivery, referred to as \textit{delivery time} (the lower, the better); (iii) the number of unsuccessful delivery attempts (the fewer, the better); and (iv) the cumulative backlog age which gives the total waiting time for all unfulfilled orders (the lower, the better).

\section{DECENTRALISED LEARNING-BASED DEPLOYMENT STRATEGY}
\label{sec:method}

\begin{figure*}[!t]
    \centering    \includegraphics[width=0.9841\linewidth]{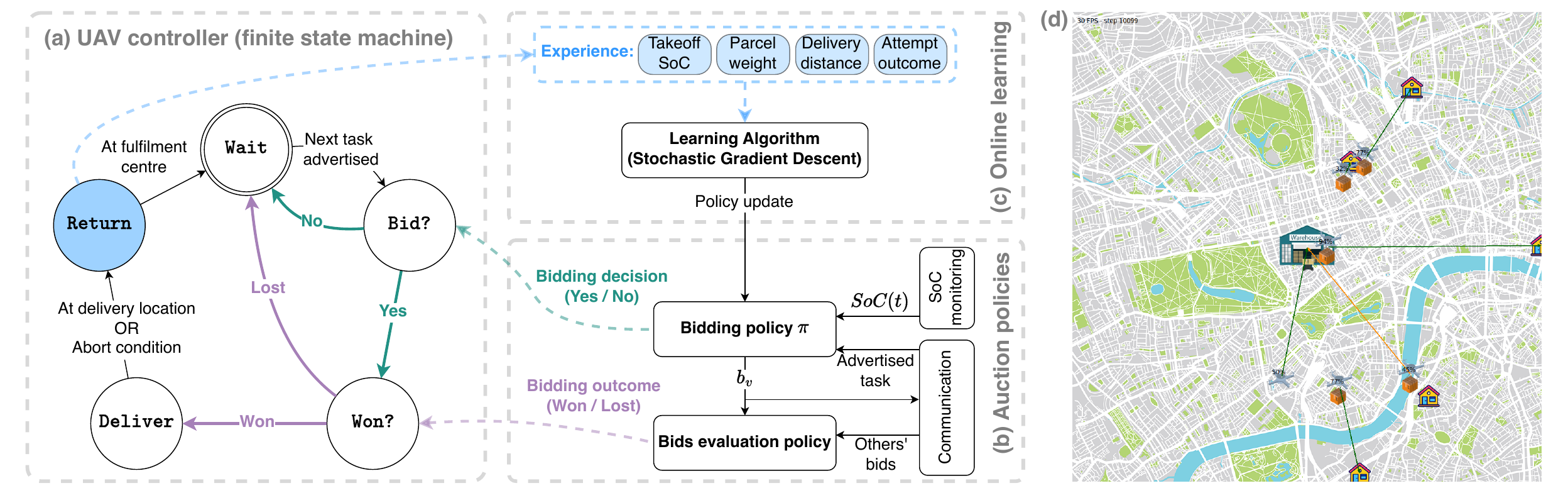}
    \Description[<Overview of the decentralised learning-based deployment strategy and evaluation environment>]{<Overview of the decentralised learning-based deployment strategy and evaluation environment>}
    \caption{Overview of decentralised learning-based deployment strategy and evaluation environment: (a) A UAV's logic is governed by a finite-state machine;
    (b) the UAV uses 
    a bidding policy to determine whether to bid (and an associated level of confidence) and a bids evaluation policy to determine whether its bid won;
    (c) upon returning from a delivery attempt, the UAV updates its bidding policy; (d) Screenshot of the purpose-built simulator showing the fulfilment centre and six drones, five of which are attempting a delivery, whereas the sixth is returning following delivery of a parcel.}
    \label{fig:overall_startegy}
\end{figure*}

We propose a decentralised deployment strategy that augments a finite-state machine controller with auction-based and online learning approaches for assigning UAVs to advertised delivery tasks. The strategy runs onboard each UAV. It comprises four components (see Figure~\ref{fig:overall_startegy}): (i) a \emph{UAV controller} defining the overall logic to transition among behavioural states;
(ii) a \emph{bidding} policy to decide for any advertised delivery task whether to place a bid and a bid value, reflecting its confidence in bidding; 
(iii) a \emph{bids evaluation} policy to determine the winner amongst the UAVs that bid for the same task; 
(iv) and an \emph{online learning algorithm} to refine the individual bidding policy such that the UAV bids only for tasks it is likely capable of completing successfully.
\subsection{UAV Controller}
We assume that all UAVs initially reside within the FC.
The UAV controller is depicted in Figure~\ref{fig:overall_startegy}a. A UAV  begins in the \(\texttt{Wait}\) state, where it awaits an announcement 
of the next delivery task
along with the task ID. A task announcement consists of a unique task id, parcel mass and delivery distance.
When a task announcement is received, the UAV transitions to the \(\texttt{Bid?}\) state. In this state, it uses its bidding policy to determine whether to bid, and a bid value, reflecting its level of confidence in the bid.
If the UAV opts to bid, it proceeds to the \(\texttt{Won?}\) state; otherwise, it returns to the \(\texttt{Wait}\) state. 
In the \(\texttt{Won?}\) state, the UAV broadcasts a tuple comprising (i) the task ID, (ii) its unique ID,  and (iii) the bid value for placing the bid. In parallel,
it records its bid and those of any other UAVs bidding for the same task. 
Subsequently, it provides all bids to its local \emph{bids evaluation} policy, which enables it to determine the outcome of the auction.
If it won, the UAV records its current state of charge, hereafter denoted as \(\text{SoC}_{\text{takeoff}}\), and transitions to the \(\texttt{Deliver}\) state. Otherwise, it returns to the \(\texttt{Wait}\) state. When in the \(\texttt{Deliver}\) state, the UAV retrieves the parcel (assumed to happen instantaneously) and flies to the delivery destination. It either (i) reaches the destination and the delivery is successful, or (ii) meets an abort condition, in which case the delivery is unsuccessful. In either case, the UAV transitions to the \(\texttt{Return}\) state. The abort condition is set such that the
UAV returns to the FC if \(\text{SoC}(t)\leq\xi \text{SoC}_{\text{takeoff}}\), that is,  \cmt{when only a fraction $\xi$ of its takeoff state of charge remains.} 
The controller is formally modelled using supervisory control theory~\cite{ramadge1987Supervisory,lopes2016Supervisory}; all models are available at \cite{EADroneDelivery_Simulator}.

\subsection{Auction Policies}
As depicted in Figure~\ref{fig:overall_startegy}b, the UAV's role in the auction-based task allocation process is implemented through two policies: one for \emph{bidding} (to decide whether to bid and the bid value)
and another for \emph{bids evaluation} (to determine whether the bid was successful).

\subsubsection{Bidding Policy}
Let UAV $i$ consider a delivery task $j \in O$ at time $t$. 
Its bidding policy, \(\pi(\cdot)\), takes as input the task's distance \(d_j\), the parcel's mass \(m_j\), and the UAV's current state of charge \(\text{SoC}_i(t)\). The features' vector is defined as \(\boldsymbol{x}^\intercal = \begin{bmatrix}
           d_{j},
           m_{j},
           \text{SoC}_i(t)
         \end{bmatrix}\).
The bidding policy outputs a binary bidding decision \(b_d\) (i.e. whether to bid or not) and a bid value \(b_v\):
\begin{equation}
    \pi(\boldsymbol{x}) = [b_d(\boldsymbol{x}), b_v(\boldsymbol{x})]
\end{equation}

\paragraph{Bidding decision} The function \(b_d(\boldsymbol{x})\) classifies input features \(\boldsymbol{x}\) into two categories: $1$ if the UAV is capable of performing the task and $0$ if not:
\begin{equation}
    b_d(\boldsymbol{x}) = 
    \begin{cases}
        1 ,&  f(\boldsymbol{x}) \geq 0\\
        0 ,& f(\boldsymbol{x}) < 0
    \end{cases}
\end{equation}
where \(f(\boldsymbol{x})\) is the decision function, in this study, a support vector machine:
\begin{equation}
   f(\boldsymbol{x}) = \boldsymbol{w}^\intercal \boldsymbol{x} + b 
   \label{eq:decision_function}
\end{equation}
where \(\boldsymbol{w}\) and \(b\) are the weight vector and bias, respectively. This decision function represents a hyperplane that separates the two categories of input features. Despite their simplicity, linear classifiers like this have demonstrated accuracy comparable to more complex nonlinear classifiers while significantly reducing training times~\cite{Yuan2012,Pang2015}. The parameter values are specific to each UAV and are updated online by the UAV after each delivery attempt.


\paragraph{Bid value} 
If a UAV decides to place a bid, a bid value has to be determined. 
It is calculated as the distance of the input features \(\boldsymbol{x}\) from the decision hyperplane \(f(\boldsymbol{x})\):
\begin{equation}
    b_v(\boldsymbol{x}) = \frac{f(\boldsymbol{x})}{\norm{\boldsymbol{w}}_2} 
\end{equation}
where \(\norm{\cdot}_2\) is the Euclidean norm. This value of \(b_v(\boldsymbol{x})\) reflects the UAV's perceived ability to complete the task, with a greater distance from the hyperplane indicating higher confidence.




\subsubsection{Bids Evaluation Policy}  

The bids evaluation policy processes the bids of all competing UAVs (including the UAV itself), returning \texttt{True} if the UAV is deemed the winner and \texttt{False} otherwise. 
We consider the following options for selecting the winner:
\begin{itemize}
\item \emph{least confident}: the UAV with the lowest bid value, that is, of lowest confidence, is the winner;
\item \emph{most confident}: the UAV with the highest bid value, that is, of highest confidence, is the winner.
\end{itemize}
In both cases, if a tie occurs, the UAV with the highest ID is selected as the winner.


\subsection{Online Learning}  
After each delivery attempt, a UAV refines its bidding policy, as expressed by decision function \(f(\boldsymbol{x})\) through continuous online learning. When attempting a delivery task \(j \in O\), it collects a labelled data point \((\boldsymbol{x}_a, y_a)\). The vector \(\boldsymbol{x}_a\) represents the features used by the UAV earlier in the bidding process, given by \(\boldsymbol{x}_a = [d_j, m_j, \text{SoC}_{\text{takeoff}}]\).
Label \(y_a\) indicates the delivery outcome: \(1\) for a successful delivery attempt and \(0\) for an aborted delivery attempt.

Upon collecting a labelled data point, the UAV updates its decision function \(f(\boldsymbol{x})\) using the Stochastic Gradient Descent (SGD) algorithm~\cite{ketkar2017stochastic}. To account for the sensitivity of SGD to the scale of input data, the features are standardised before updating:
\begin{equation}
    \boldsymbol{\Tilde{x}}_a = \dfrac{\boldsymbol{x}_a - \boldsymbol{\mu}}{\boldsymbol{\sigma}}
    \label{eq:standard_scaler}
\end{equation}
where \(\boldsymbol{\Tilde{x}}_a\) represents the standardised input features, \(\boldsymbol{x}_a\) the original input features, and \(\boldsymbol{\mu}\) and \(\boldsymbol{\sigma}\) denote the mean and standard deviation of the input features vector, respectively.

The decision function updates are carried out with the objective of minimising the regularised training error:
\begin{equation}
    E(w, b) = L(y_a, f(\boldsymbol{\Tilde{x}}_a)) + \alpha R(w)
\end{equation}
where \(L\) is the loss function measuring the (mis)fit of the decision function, and \(R\) is a regularisation term. The hyperparameter \(\alpha > 0\) controls the strength of the regularisation.

In our implementation, we employ a modified Huber loss function for its proven ability to enable fast and robust learning~\cite{Kim2011,guo2021modified}:
\begin{align}
&L(y_a, f(\boldsymbol{\Tilde{x}}_a)) = \notag \\
&\begin{cases} 
\max\Big(0, 1 - (2y_a-1) f(\boldsymbol{\Tilde{x}}_a)\Big)^2, & \text{if } (2y_a - 1) f(\boldsymbol{\Tilde{x}}_a) \geq -1, \\
-4 (2y_a - 1) f(\boldsymbol{\Tilde{x}}_a), & \text{otherwise} 
\end{cases}
\end{align}
and an \(\ell_2\) norm regularisation term:
\begin{equation}
R(\boldsymbol{w}) = \frac{\norm{\boldsymbol{w}}_2}{2}
\end{equation}

Using SGD, the parameters of the decision function are updated iteratively as follows:
\begin{equation}
 \begin{cases}
        & \boldsymbol{w} \leftarrow \boldsymbol{w} - \eta \left[\frac{\partial L(y_a, f(\boldsymbol{\Tilde{x}}_a))}{\partial \boldsymbol{w}} + \alpha \frac{\partial R(\boldsymbol{w})}{\partial \boldsymbol{w}} \right], \\ \\
        & b \leftarrow b - \eta \frac{\partial L(y_a, f(\boldsymbol{\Tilde{x}}_a))}{\partial b}
\end{cases}
\end{equation}
where \(\eta\) is the learning rate. To improve convergence and stability, \(\eta\) progressively decays over the optimisation steps in the individual SGD runs, following the heuristic approach proposed by~\cite{bottou2012stochastic}.

The decision function is initially trained based on two assumed labelled data points: (i) \(([d_{\text{min}},m_{\text{min}},100.0],1)\), suggesting that a fully charged UAV, that is, $\text{SoC}_{\text{takeoff}}=100$, would successfully deliver a parcel of minimum mass \(m_{\text{min}}\) for the minimum delivery distance \(d_{\text{min}}\); (ii) \(([d_{\text{max}},m_{\text{max}},0.0],0)\), suggesting that a completely fully discharged UAV, that is, $\text{SoC}_{\text{takeoff}}=0$, cannot deliver a parcel of maximum mass \(m_{\text{max}}\) for the maximum delivery distance \(d_{\text{max}}\).

\section{Multi-Agent Simulations}
\label{sec:experimental-setup}


To enable a comprehensive performance evaluation of strategies in long-term deployment scenarios---with up to ca. 2.4 million auctions over 8 weeks of simulated time per trial, we developed an ultra-fast Python-based multi-agent simulator, the source code of which is available online at~\cite{EADroneDelivery_Simulator}. A screenshot of the simulator is shown in Figure~\ref{fig:overall_startegy}d. To implement the learning algorithm, the widely-used scikit-learn library~\cite{scikit-learn} is used.

\subsection{UAV Energy Models}
\label{sec:energy_models}
The simulator incorporates realistic UAV energy models for both battery charging and consumption. 
\subsubsection*{Charging model}
The charging model is based on the following assumptions: (i) The charging rate decreases as the battery approaches maximum capacity due to increasing internal resistance; (ii) the time to fully charge
is independent 
of the battery’s SoH---although batteries with a lower SoH have reduced capacity, they are known to charge less efficiently~\cite{Gong2015, Lee2021, Arof2023}.

When charging, the UAV's state of charge, $\text{SoC}(t)\in[0,100]$, increases at the following rate:
\begin{equation}
    \lambda_{\text{charge}}(t) = \dfrac{100 - \text{SoC}(t)}{3600} \dfrac{\gamma_c P_c}{C_{\text{theoretical}}}
\end{equation}
where \(P_c\) and \(\gamma_c\) denote the charger’s power (in \(W\)) and efficiency. \(C_{\text{theoretical}}\) refers to the UAV's theoretical battery capacity (in \(Wh\)).

\subsubsection*{Consumption model}
We use the model proposed in~\cite{dorling2016vehicle} to compute the energy consumption (in \(Ws\)) per second for a UAV flying with a payload mass $m_p$ at a constant speed \(v_a\):
\begin{equation}
    C_{ps}(m_p) = \dfrac{[g (m_f + m_b + m_p)]^{3/2}}{3600\sqrt{2 n_r\rho \zeta}}
\end{equation}
where $m_f$ and $m_b$ are the UAV frame and battery masses, respectively. $n_r$ is the number of rotors on the UAV, $\zeta$ is the area of the spinning blade disc of one rotor, $g$ is the gravitational constant (in $m/s^2$), and $\rho$ is the air density at $15^{\circ}\mathrm{C}$. \cmt{It is important to note that the UAV does not have access to this energy consumption model; the model is used solely for simulation purposes.}

When delivering, the UAV's state of charge, $\text{SoC}(t)\in[0,100]$, decreases at the following rate:
\begin{equation}
    \lambda_{\text{delivery}}(m_p) = -100\dfrac{C_{ps}(m_p)}{C_{\text{theoretical}}\cdot \text{SoH}}
\end{equation}

\subsection{Orders Arrival and Processing}
In our simulations, the FC is responsible for receiving orders, preparing parcels, and announcing the corresponding delivery tasks to available UAVs.

\paragraph{Order Arrival}
Orders arrive stochastically, with an expected inter-arrival time \(\overline{\tau}\). The arrival time of the next order is randomly drawn from an exponential distribution:
\begin{equation}
    t_{j+1} = t_{j} + Z_j\quad \text{where } Z_j \sim \text{Exp}\left(\frac{1}{\overline{\tau}}\right)\quad \text{with } t_1 = 0.
\end{equation}

\paragraph{Order Processing}
Orders arriving at the FC are stored in a queue which sorts them by their arrival times from the earliest to the latest placed orders. The advertising of tasks, extracted from these orders, starts with the first order in the queue. It adheres to the following rules:
(i) for as long as no UAV resides at the FC, task advertisement is suspended;
(ii) 
if a task is advertised but no UAV places a bid, the FC proceeds with the next unallocated order in the queue, if any; 
(iii) if a task is advertised,  and at least one UAV places a bid, the order is considered allocated; the FC continues with the earliest unallocated order in the queue (i.e., considering again the front of the queue); 
(iv) after an aborted delivery attempt, the corresponding order is considered unallocated again; it remains in the queue according to the original arrival time of the order. 

    %






\begin{table}
\caption{Simulation variables.}
\centering
\resizebox{\columnwidth}{!}{%
\begin{tabular}{ccccc}
\toprule
\textbf{Variables}                  & \textbf{Denotation}        & \textbf{Description}                         & \textbf{Value(s)} & \textbf{Unit} \\ 
\midrule
\multirow{4}{*}{UAV}                & \(m_f\)                    & Frame mass                                   & 10               & kg            \\ 
                                    & \(n_r\)                    & Number of rotors                             & 8               & ---           \\ 
                                    & \(\zeta\)                  & Area of the spinning blade disc of one rotor & 0.27               & \(m^2\)       \\ 
                                    & \(v_a\)                    & Nominal speed                                & 10               & \(m/s\)       \\
\midrule
\multirow{3}{*}{Battery}            & \(m_b\)                    & Mass                                 & 10               & \(kg\)        \\ 
                                    & \(C_{\text{theoretical}}\) & Theoretical capacity                 & 800               & \(Wh\)        \\
                                    & \(\text{SoH}\)         & State of Health                       & \(\mathcal{U}(0.5,1.0)\)               & ---           \\ 
\midrule
\multirow{2}{*}{Charger}            & \(P_c\)                      & Power                                        & 100               & \(W\)         \\  
                                    & \(\gamma_c \)                   & Efficiency                                   & 95\%               & ---           \\ 
\midrule
\multirow{1}{*}{Swarm}              & \(S\)                      & Swarm size                                   & 25               & ---           \\ 
\midrule
\multirow{4}{*}{Orders}             & \(d_j\)           & Delivery distance                    & \(\mathcal{U}(1000,6000)\)               & \(m\)         \\ 
                                    & \(m_j\)         & Parcel mass                          & \(\mathcal{U}(0.5,5.0)\)              & \(kg\)        \\
                                    & \(\overline{\tau}\)         & Inter-arrival time                          & \(\{15,20,25,30,35,40\}\)              & \(min\)        \\
                                    & \(T\)                      & Operation period                              & 8                & weeks           \\

\midrule
\multirow{2}{*}{Other}              & \(g\)                      & Gravitational acceleration constant          & 9.81               & \(m/s^2\)     \\ 
                                    & \(\rho\)                   & Air density at 15 C\(^{\circ}\)              & 1.225               & \(kg/m^3\)    \\ 
\bottomrule
\end{tabular}%
}
\label{tab:parameter-table}
\end{table}

\section{RESULTS}
\label{sec:results}
In this section, we present a comprehensive performance evaluation of the proposed learning-based deployment strategy.
Table~\ref{tab:parameter-table} provides an overview of the simulation parameters. 
Unless otherwise stated, we simulate fleets of \(S=25\) UAVs over an operational period of \(T=8\) weeks. 
We consider order inter-arrival times of \(\overline{\tau} \in \{15, 20, 25, 30, 35, 40\}\) minutes. The order parameters are chosen at random using uniform distributions, \(d_j \sim \mathcal{U}(1000, 6000)\) metres and \(m_j \sim \mathcal{U}(0.5, 5.0)\) kilograms. If meeting the aforementioned conditions, orders are advertised one at a time, every 2\,s.
Table~\ref{tab:parameter-table} provides as well the parameters used to simulate the UAV, including its battery, and the charger. For each UAV, the battery health is chosen randomly using a uniform distribution, \(\text{SoH} \sim \mathcal{U}(0.5, 1.0)\). Assuming $\text{SoH}=1$ and using the energy models described in Section~\ref{sec:energy_models}, the UAV can fly for 28.75 min while transporting a parcel of maximum weight. It requires 505 min to fully charge its battery once entirely depleted. Hence, its maximum duty is 5.4\%, which is similar to the duty cycle of the commercial delivery drone DJI FlyCart 30 (6.97\% with the single-battery configuration).

UAVs are set to abort delivery attempts \cmt{upon reaching} $\xi=0.5$ of their takeoff state of charge, assuming equal energy for the return journey. \cmt{Lower \(\xi\) values delay returns but risk UAV loss (as confirmed via separate simulations which are not included in the paper), making them impractical.}   

For the learning algorithm, we set both the regularisation strength parameter \(\alpha\) and the initial learning rate \(\eta\) to \(0.01\). Given that delivery distance \(d_j\) is sampled from \(\mathcal{U}(1000, 6000)\), parcel mass \(m_j\) is drawn from \(\mathcal{U}(0.5, 5.0)\), and robots can have a SoC ranging from \(0\) to \(100\), we set the mean vector and standard deviation vector in Equation~\ref{eq:standard_scaler} to
$\boldsymbol{\mu} = (3500, 2.75, 50)$ and
$\boldsymbol{\sigma} = (1443.38, 1.298, 28.87)$.

Each experimental condition is repeated over 20 random runs, leveraging an Intel Xeon Platinum 8358 CPU core in a High-Perfor\-mance Computing (HPC) cluster running Python 3.10.12.

\subsection{Effect of Winner Selection Rule}

\begin{figure*}[!t]
    \centering
    \includegraphics[width=1.0\linewidth]{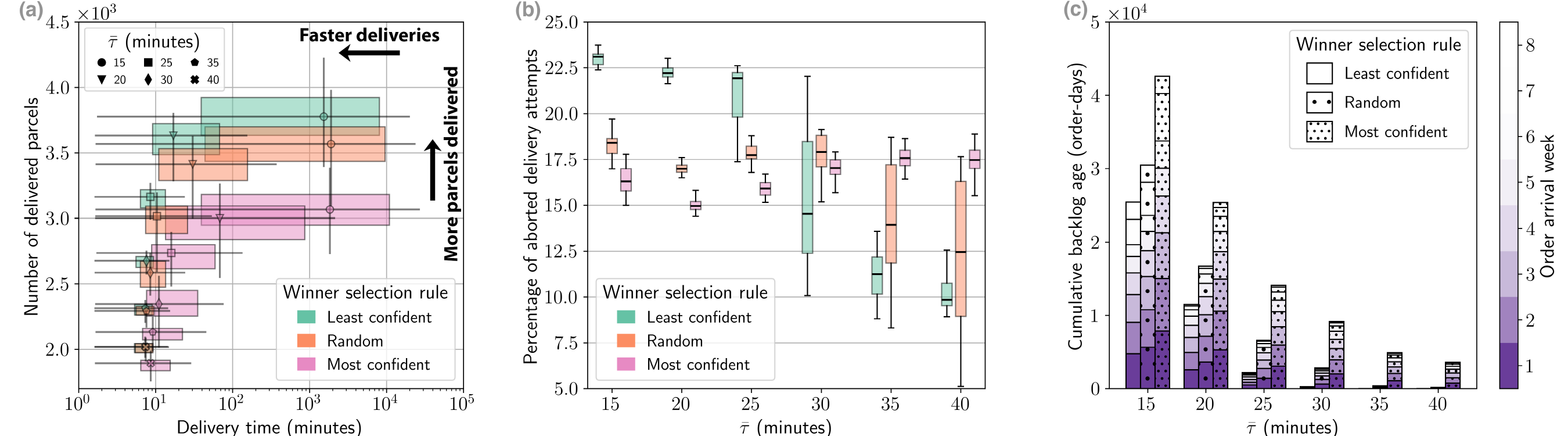}
    \Description[<Comparing different variants>]{<Comparing different variants>}
    \caption{Performance of the learning-based deployment strategy for different winner selection rules: \emph{Least confident} (green), \emph{Most confident} (pink), and \emph{Random} (orange). Metrics used are  (a) number of delivered parcels and delivery time, (b) percentage of aborted delivery attempts, and (c) cumulative backlog age (segmented by order arrival weeks).}
    \label{fig:winner_selection_rules_comparison}
\end{figure*}


In Figure~\ref{fig:winner_selection_rules_comparison}, we evaluate our learning-based deployment strategy under three winner selection rules: \emph{Least confident} (green), \emph{Most confident} (pink), and \emph{Random} (orange). \emph{Random} (orange) effectively chooses each bidder with equal probability.\footnote{To achieve this distribution in a decentralised way, each bidder randomly samples their bid value, and once all bids have been broadcast, the highest bidder wins.}
Figure~\ref{fig:winner_selection_rules_comparison}a compares the winner selection rules in terms of delivery time (x-axis) and number of delivered parcels (y-axis) for different inter-arrival intervals. The distributions (over 20 runs) are visualised using a 2D box plot for each combination of winner selection rule and inter-arrival interval. 
Counter-intuitively, the \emph{Least confident} rule outperforms the other two, consistently delivering more parcels, and in shorter times across all inter-arrival intervals. This can possibly be attributed to a more demand-oriented workload allocation where less capable UAVs end up fulfilling simpler tasks, preserving the more capable UAVs for more demanding tasks that may arise in the future. Moreover, when the least confident bidder is employed, it is possible that the UAVs learn to set their confidence levels conservatively. The \emph{Most confident} rule leads to fewer deliveries and longer delivery times, whereas the \emph{Random} strategy falls between the two.  

Figure~\ref{fig:winner_selection_rules_comparison}b illustrates the percentage of aborted delivery attempts (y-axis) for various expected inter-arrival times (\(\overline{\tau}\)) (x-axis) across the three winner selection rules. 
For shorter inter-arrival times (\(\overline{\tau} \leq 25\) minutes), the \emph{Least confident} rule results in a higher percentage of aborted deliveries. This could be because less confident UAVs are more likely to operate close to their capability limit.
For longer inter-arrival times (\(\overline{\tau} > 25\)), the \emph{Least confident} rule exhibits consistently the fewest aborted attempts of the three rules.
The performance of the \emph{Random} rule notably improves as well. By contrast, the percentage of aborted delivery attempts does not decrease for the \emph{Most confident} rule as the inter-arrival time increases.

Figure~\ref{fig:winner_selection_rules_comparison}c shows a bar plot of the cumulative backlog age at the end of the 8-week operation period for different inter-arrival times (\(\overline{\tau}\)) and the different winner selection rules. Each bar further breaks down the orders by their arrival week. As \(\overline{\tau}\) increases from 15 to 40 minutes, all selection rules show a reduction in the cumulative backlog age, suggesting that less frequent order arrivals lead to lower backlogs. The \emph{Least confident} rule consistently yields the lowest cumulative backlog age (for all inter-arrival times). This indicates better queue management and backlog reduction, especially as the inter-arrival time increases, being able to process almost all orders for (\(\overline{\tau} \geq 30\)). By contrast, the \emph{Most confident} rule results in the highest cumulative backlog age, implying a less efficient resource use. Note that depending on the inter-arrival time, 252--672 orders are expected to arrive per week. 

\subsection{Analysis of Decision Accuracies}


We evaluate the decision-making accuracy of the learned bidding policies at the end of each week of the eight-week operation period. All three selection rules are considered. For each rule, 500 UAVs are evaluated (i.e., 25 UAVs per run, and 20 runs in total). 
The inter-arrival time is \(\overline{\tau} = 15\) minutes.

We computed the decisions made by the bidding policies (i.e., whether to bid or not) using 1000 randomly generated input feature vectors \(\boldsymbol{x}^\intercal = \begin{bmatrix}
           d,
           m,
           \text{SoC}
         \end{bmatrix}\), 
where \(d \sim \mathcal{U}(1000, 6000)\), \(m \sim \mathcal{U}(0.5, 5.0)\), and \(\text{SoC} \sim \mathcal{U}(0, 100)\). We then assessed the correctness of these decisions by comparing them to the ground truth. The ground truth uses the UAVs' true battery capacity and the simulated energy consumption model (both of which are unknown to the UAV) to determine whether the UAV can complete the task. A decision is considered correct only if the UAV chooses to bid for a task it is capable of completing, or if it chooses not to bid for a task it is incapable of completing. 

Figure~\ref{fig:why_least_confident_better}a displays the mean decision accuracy (y-axis) over time (x-axis) for three winner selection strategies: \emph{Least confident} (green), \emph{Most confident} (pink), and \emph{Random} (orange). The shaded areas around the line plots represent the 95\% confidence intervals for the mean decision accuracy. The \emph{Least confident} rule exhibited significantly better decision accuracy, reaching about 97\% after eight weeks. This could be attributed to these UAVs tending to push their capabilities more frequently, leading to more refined decision-making over time. In contrast, the \emph{Most confident} rule showed slower progress, remaining below 85\% after eight weeks. Many of these UAVs have fewer opportunities to learn their limits. 

\subsection{Analysis of Energy Constraints of Winning Bidder}

\begin{figure*}[!t]
    \centering
    \includegraphics[width=1.0\linewidth]{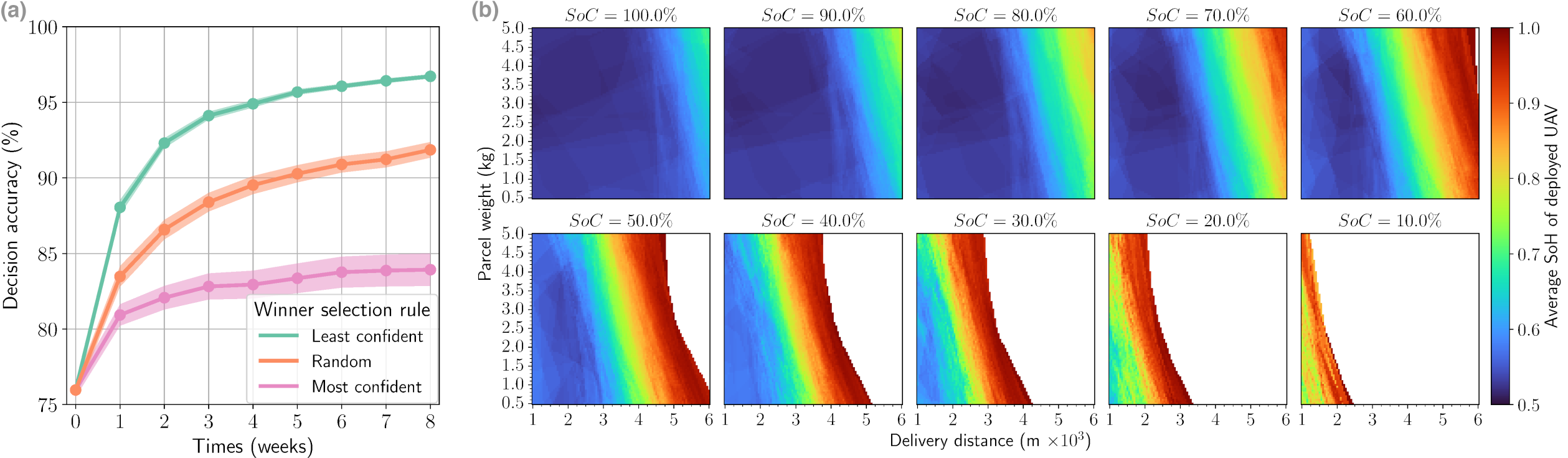}
    \Description[<Why least confidence is better>]{<Why least confidence is better>}
    \caption{(a) Decision accuracy over time for the three winner selection strategies: \emph{Least confident} (green), \emph{Most confident} (pink), and \emph{Random} (orange). 
    (b) SoH of UAVs deployed by the \emph{Least confident} winner selection rule for various tasks and SoC values.
    }
    \label{fig:why_least_confident_better}
\end{figure*}


We examine the extent to which a UAV's state of health (indirectly) impacts its bidding policy. Recall that the UAV is unaware of its state of health. We test the bidding policies that are obtained at the end of the 8-week operation period for UAVs that used the \emph{Least confident} winner selection rule during learning. We then record the bid value for every UAV, assuming a \(\text{SoC} \in \{10, 20, \dots, 100\}\), a delivery distance \(d_j \in \{1000, 1050, \dots, 5000\}\) metres and mass \(m_j \in \{0.5, 0.6, \dots, 5\}\) kilograms. For each combination of SoC, distance, and mass, we record the SoH of the winning UAV (if any bids are placed). The reported values are the average SoH across 20 runs. 

The results are depicted in Figure~\ref{fig:why_least_confident_better}b.
Each subplot corresponds to a specific SoC.
The colour gradient, from blue to red, represents the average SoH of the winning UAVs (the colour being white if no bids were placed).
When SoC is 100\%, the SoH of the deployed UAVs ranges from 0.5 to around 0.7, with more difficult tasks (longer distances and heavier parcels) corresponding to higher SoH values. This suggests that the \emph{Least confident} strategy prioritizes deploying the least capable UAVs that can still complete the task. As SoC decreases, the range of SoH values expands. For instance, when SoC is 60\%, the SoH of the deployed UAVs ranges from 0.5 to 1.0, with more difficult tasks leading to the deployment of UAVs with higher SoH. However, as SoC continues to drop, we observe that UAVs are no longer deployed for harder tasks (upper right region of the heatmaps). This occurs because, at low SoC levels, even the most capable UAVs (those with high SoH) are unable to handle these challenging deliveries. When SoC reaches very low levels (e.g., 10\%), UAVs are only deployed for the simplest tasks, and the SoH values are restricted to higher ranges (above 0.7). 
At such low energy levels, only the most capable UAVs are selected for the remaining simpler tasks, ensuring that the system operates within the UAVs' remaining capacity.

This result highlights that the \emph{Least confident} winner selection rule succeeds in selecting UAVs that are just capable of performing the given tasks, thereby preserving the more capable UAVs for more demanding tasks. Moreover, despite the UAV being unaware of its own state of health, the behaviour it learns is clearly correlated with it. This shows that the overall decision-making process has learned an implicit awareness of the individual energy constraints of the UAV.


\subsection{Comparison Against Threshold-based Deployment Strategy}

In the following, the learning-based deployment strategy is only considered in conjunction with the \emph{Least confident} winner selection rule.
We benchmark this strategy against a traditional \textit{threshold-based} deployment strategy. In this baseline approach, UAVs are only deployed if their SoC exceeds a predefined threshold. This strategy operates using the same auction-based mechanism for task allocation as the learning-based approach (see Section~\ref{sec:method}), but UAVs bid only when their SoC is above the threshold, and their bid value corresponds to the current SoC. We tested this threshold-based strategy with SoC thresholds of $\{50, 60, 70, 80, 90, 100\} \%$
and found that the threshold-based strategy achieved the highest number of deliveries and, on average, the shortest delivery times when the UAVs deployed a SoC threshold of $80\%$. 
Note that the learning-based approach starts without such prior tuning, as the UAVs need to progressively refine their bidding strategies. 

\begin{figure*}[!th]
    \centering
    \includegraphics[width=1.0\linewidth]{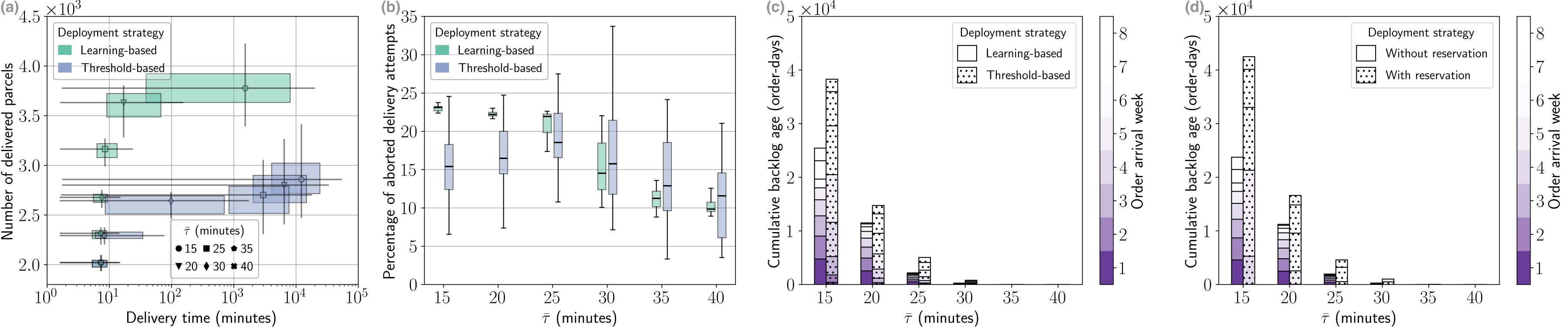}
    \Description[<Comparing the proposed learning-based deployment strategy against the baseline approach>]{<Comparing the proposed deployment strategy against the baseline approach>}
    \caption{Comparing the learning-based deployment strategy against the threshold-based deployment strategy (a--c); (a) The learning-based strategy outperforms the threshold-based strategy in terms of the number of delivered parcels and delivery time, (b) results in a higher number of failed delivery attempts, (c) but produces less backlog. \cmt{(d) Cumulative backlog age after eight weeks for the learning-based strategy with and without reservations.}}
    \label{fig:results:learning_vs_fixed}
\end{figure*}

Figure~\ref{fig:results:learning_vs_fixed}a presents the results in
terms of the number of delivered parcels (y-axis) and delivery time (x-axis), across various inter-arrival times (\(\overline{\tau}\)) from 15 to 40 min. At the lowest expected inter-arrival time of 40 min, both strategies perform equally well regarding both criteria. As the expected inter-arrival time increases, both strategies manage to deliver more parcels, but the performance gap widens. The \emph{Learning-based} strategy consistently outperforms the \emph{Threshold-based} strategy in both delivery volume and delivery time. 
For example, at \(\overline{\tau} = 20\) minutes, the \emph{Learning-based} strategy delivers a median of approximately $3600$ parcels, with a median delivery time of $17$ min, whereas the \emph{Threshold-based} strategy delivers approximately $2800$ parcels, with a median delivery time exceeding $6500$ minutes. This stark difference may be attributed to the adaptive nature of the \emph{Learning-based} approach, which allows UAVs to bid and be deployed at any SoC level, as long as considered capable of completing the given task, while the \emph{Threshold-based} strategy waits for UAVs to reach the required SoC level, regardless of the task difficulty. 


In Figure~\ref{fig:results:learning_vs_fixed}b, we compare two strategies in terms of aborted delivery attempts. For low inter-arrival intervals (\(\overline{\tau} \leq 25\)), the \textit{Learning-based} strategy exhibits a higher number of aborted delivery attempts than the \textit{Threshold-based} strategy. This is because the \textit{Learning-based} strategy allows UAVs to attempt deliveries at lower SoC levels. However, despite the increased 
aborted attempts, the strategy achieves lower delivery times.


In Figure~\ref{fig:results:learning_vs_fixed}c, we observe that the \emph{Learning-based} strategy consistently results in a lower cumulative backlog age compared to the \emph{Threshold-based} strategy, but it is less good at prioritising orders from earlier weeks. This could be because the \textit{Learning-based} strategy's responsiveness encourages UAVs to be deployed as soon they consider themselves capable of fulfilling the order, leading to faster task completion and less accumulation of pending orders. On the other hand, the \textit{Threshold-based} strategy leads to delayed deployments, which exacerbates task backlog, especially when tasks arrive in rapid succession.


In summary, the \textit{Learning-based} strategy not only improves delivery throughput, but also reduces delivery time and backlog, despite having a higher rate of aborted attempts compared to the \textit{Threshold-based} strategy. \cmt{These findings also hold for higher \( \xi \) values (\(\xi = 0.55, 0.6, 0.65, 0.7\)), which provide an increased safety margin for returns---critical for example when accounting for changing wind direction---albeit at the expense of fewer deliveries for both strategies (see Figure S1 in the supplementary material~\cite{supplementary_material}).}

\cmt{The impact of fleet size, originally $S=25$, was also examined (see Figure S2 in the supplementary material~\cite{supplementary_material}). For a fixed arrival time, reducing the fleet size amplifies the advantage of the \textit{Learning-based} strategy, whereas increasing it diminishes the advantage. 
}

\subsection{Forecasting Using Learned Bidding Policies}

In this section, we explore
an enhancement enabling UAVs with insufficient SoC to commit to fulfilling current delivery tasks at a later time, once they have accumulated enough charge. UAVs that choose not to bid for immediate delivery can submit a \textit{reservation bid}. Each UAV uses its learned decision function \( f(\boldsymbol{x}) \) and the expected charging rate to determine whether to place a reservation bid and to calculate the appropriate bid value. A reservation bid is made if the UAV forecasts that it will be able to perform the task in the future. The bid value represents the forecasted time required to reach the sufficient charge and begin addressing the task. A UAV wins the reservation auction if no other UAV bids to address the task immediately, and its bid value is the lowest. 


We evaluated this forecasting variant in simulation over an 8-week operational period. To assess the effect of forecasting, we used UAVs that had previously learned and tuned their decision functions over an earlier 8-week period. Figure~\ref{fig:results:learning_vs_fixed}d illustrates the impact of forecasting on task backlog. The results show a significant reduction in the number of pending orders from earlier weeks when forecasting is enabled compared to when it is not. However, more backlogs will be created for later weeks. By allowing UAVs to commit to future tasks, the forecasting mechanism helps prioritise earlier orders, but increases backlog.


\section{CONCLUSION}
\label{sec:conclusion}


We considered an on-demand delivery scenario, where fleets of UAVs address orders that arrive stochastically at a fulfilment centre (FC). The UAVs have heterogeneous capabilities, differing in battery health, which determines the maximum energy they can store. They are neither aware of their battery health---and hence their true energy capacity---nor of their energy consumption models. To address this, we proposed a decentralised deployment strategy that combines auction-based task allocation with online learning. 
Each UAV independently decides whether to bid for an order based on its current state of charge, the delivery distance, and parcel mass.  
The UAV continuously refines its bidding policy to become consistent with its individual capability for fulfilling orders.

Through an extensive set of simulations, we demonstrated that our learning-based deployment strategy outperforms a traditional threshold-based approach, which only deploys UAVs when their state of charge exceeds a predefined amount. The learning-based strategy consistently delivered more parcels and in shorter times, proving particularly effective in scenarios with frequent task arrivals. Our analysis revealed that deploying the least confident bidders led to more efficient workload distribution and better overall delivery performance. The learning-based strategy made it possible for UAVs to adapt their decision-making to their individual energy constraints, including the state of charge and unknown state of health, in addition to being responsive to the task parameters. The strategy was subsequently extended to enable forecasting via the use of the learned capability models. This enabled UAVs with insufficient charge levels to commit to fulfilling orders at specific future times. This also highlights the flexibility of our approach in handling real-world delivery challenges, such as fluctuating demands and task prioritisation.

\cmt{
Although this paper considered fleets of UAVs with varying battery health, the proposed deployment strategy learns general capability models, allowing each UAV to identify the types of orders it likely fulfils. These models should be able to capture other device-specific factors impacting capability, such as hardware wear and tear (e.g., propeller degradation).

We focused on a decentralised implementation of the learning-based deployment strategy as it offers scalability, reduces communication overhead, and could be applied in scenarios involving multiple FCs. However, a centralised implementation could be realised, where the FC stores and updates the individual capability models and allocates delivery tasks while monitoring the charge levels of the UAVs.
}

\cmt{Future work will extend the proposed strategy to 
incorporate environmental factors such as wind speed and direction into the decision-making process.
It will consider scenarios where UAVs handle multiple parcels, which requires them to determine whether to wait for additional orders or depart based on their bidding confidence.} To accommodate these complexities, more sophisticated policy architectures, such as deep neural networks, may be required.
 

\begin{acks}
This document is issued within the frame and for the purpose of the \href{https://openswarm.eu/}{OpenSwarm} project. This project has received funding from the European Union's Horizon Europe Framework Programme under Grant Agreement No. 101093046. Views and opinions expressed are however those of the author(s) only and the European Commission is not responsible for any use that may be made of the information it contains.
\end{acks}


\balance
\bibliographystyle{ACM-Reference-Format} 
\bibliography{references}


\end{document}